\documentclass{article}

\usepackage[final]{neurips_2024}

\usepackage[utf8]{inputenc} 
\usepackage[T1]{fontenc}    
\usepackage{hyperref}       
\hypersetup{colorlinks = true, linkcolor = brown,
            urlcolor  = gray,
            citecolor = brown,
            anchorcolor = brown}
\usepackage{url}            
\usepackage{booktabs}       
\usepackage{amsfonts}       
\usepackage{nicefrac}       
\usepackage{microtype}      
\usepackage{xcolor}         

\usepackage{graphicx}

\usepackage{algorithm}
\usepackage[noend]{algpseudocode}
\usepackage{algorithmicx}

\usepackage{amsmath}
\usepackage{amssymb}
\usepackage{mathtools}
\usepackage{amsthm}

\usepackage{bm}
\usepackage[british,english,american]{babel}
\usepackage{enumitem}
\usepackage{lipsum}
\usepackage{multirow}
\usepackage{colortbl}
\usepackage{wrapfig}
\usepackage{subcaption}

\definecolor{deemph}{gray}{0.6}
\newcommand{\gc}[1]{\textcolor{deemph}{#1}}

\newcommand{\tablestyle}[2]{\setlength{\tabcolsep}{#1}\renewcommand{\arraystretch}{#2}\centering\footnotesize}

\definecolor{inacc}{rgb}{0.82, 0.82, 0.82}
\definecolor{pix}{rgb}{0.94, 0.87, 0.8}
\definecolor{ins}{rgb}{0.74, 0.83, 0.9}
\definecolor{bd}{rgb}{0., 1., 1.}
\newcommand{\gr}{\rowcolor[gray]{.95}}

\definecolor{convcolor}{HTML}{412F8A}
\definecolor{vitcolor}{HTML}{fc8e62}

\newcommand{\pixcolor}[1]{\textcolor{convcolor}{#1}}
\newcommand{\inscolor}[1]{\textcolor{vitcolor}{#1}}
\newcommand{\insb}{\inscolor{$\mathbf{\circ}$\,}}
\newcommand{\pixb}{\pixcolor{$\bullet$\,}}

\newcommand{\up}[1]{\textcolor[rgb]{0,0.75,0.25}{$\uparrow${#1}}}
\newcommand{\best}[1]{\textbf{#1}}
\newcommand{\bt}[1]{\textcolor{black}{#1}}

\newcommand{\I}{\mathbf{I}}

\title{Segment Any Change}

\author{%
Zhuo~Zheng$^{1}$,~Yanfei~Zhong$^{2}$\thanks{Corresponding authors: Stefano Ermon, Yanfei Zhong}~,~Liangpei~Zhang$^{2}$,~Stefano~Ermon$^{1}$\footnotemark[1]\\
  $^{1}$Stanford University\\
  $^{2}$Wuhan University \\
  \texttt{zhuozheng@cs.stanford.edu} \\
}

\begin{document}

\maketitle

\begin{abstract}
Visual foundation models have achieved remarkable results in zero-shot image classification and segmentation, but zero-shot change detection remains an open problem.
In this paper, we propose the segment any change models (AnyChange), a new type of change detection model that supports zero-shot prediction and generalization on unseen change types and data distributions.
AnyChange is built on the segment anything model (SAM) via our training-free adaptation method, bitemporal latent matching.
By revealing and exploiting 
intra-image and inter-image semantic similarities in SAM's latent space, bitemporal latent matching endows SAM with zero-shot change detection capabilities in a training-free way. 
We also propose a point query mechanism to
enable AnyChange's zero-shot object-centric change detection capability.
We perform extensive experiments to confirm the 
effectiveness of AnyChange for zero-shot change detection.
AnyChange sets a new record on the SECOND benchmark for unsupervised change detection, exceeding the previous SOTA by up to 4.4\% F$_1$ score, and achieving comparable accuracy with negligible manual annotations (1 pixel per image) for supervised change detection.
Code is available at \url{https://github.com/Z-Zheng/pytorch-change-models}.
\end{abstract}

\section{Introduction}
\label{intro}
The Earth's surface undergoes constant changes over time due to natural processes and human activities.
Some of the dynamic processes driving these changes (e.g., natural disasters, deforestation, and urbanization) 
have huge impact on climate, environment, and human life \citep{zhu2022remote}.
Capturing these global changes via remote sensing and machine learning is a crucial step in many sustainability disciplines \citep{yeh2020using,burke2021using}.

Deep change detection models have yielded impressive results via large-scale pre-training \citep{seco,wang2022empirical,mall2023change,changen} and architecture improvements \citep{chen2021remote,zheng2022changemask}.
However, their capabilities depend on training data and are limited to specific application scenarios.
These models cannot generalize to new change types and data distributions (e.g., new geographic areas)
beyond those seen during training.

This desired level of generalization on unseen change types and data distributions requires change detection models with zero-shot prediction capabilities.
However, the concept of \textit{zero-shot change detection} has not been explored so far in the literature.
While we are in the era of ``foundation models'' \citep{fm} and have witnessed the emergence of large language models (LLMs) and vision foundation models (VFMs) (e.g., CLIP \citep{clip} and Segment Anything Model (SAM) \citep{sam}) with strong zero-shot prediction and generalization capabilities via prompt engineering, 
zero-shot change detection is still an open problem.

\begin{figure*}[ht]
  \centering
   \includegraphics[width=1.\linewidth]{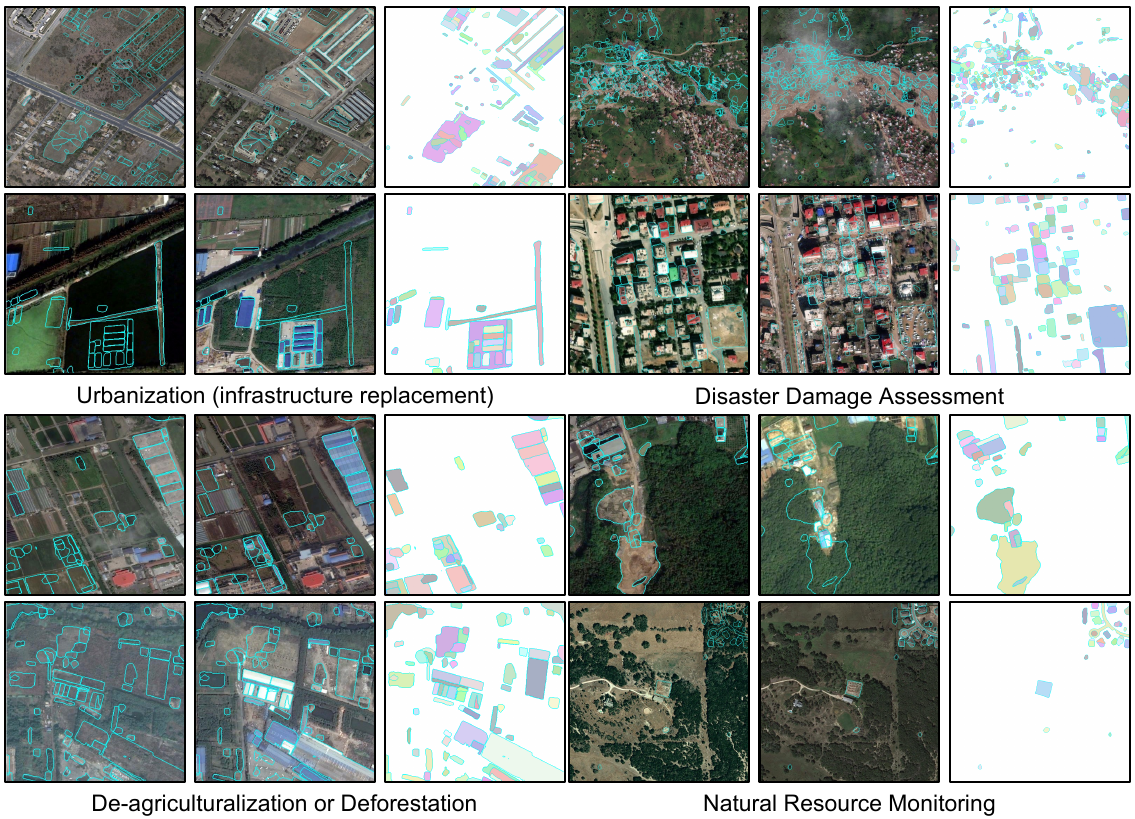}
   \vspace{-1em}
   \caption{\textbf{Zero-Shot Change Detection with AnyChange} on a wide range of application scenarios in geoscience. 
   Each subfigure presents the pre-event image, the post-event image, and their change instance masks in order. 
   The boundary of each change instance mask is rendered by \textcolor{bd}{cyan}, and meanwhile, these change masks are also drawn on pre/post-event images to show more clearly where the change occurred.
   The color of each change mask is used to distinguish between different instances.
   }
   \label{fig:teaser}
   \vspace{-1.3em}
\end{figure*}

To close this gap, we present \textit{Segment Any Change}, the first change detection model with zero-shot generalization on unseen change types and data distributions.
Our approach builds on SAM, the first promptable image segmentation model, which has shown extraordinary zero-shot generalization on object types and data distributions.
While SAM is extremely capable, it is non-trivial to adapt SAM to change detection and maintain its zero-shot generalization and promptability due to the extreme data collection cost of large-scale 
change detection labels that would be required to enable promptable training as in the original SAM.

To resolve this problem, we propose a training-free method, \textit{bitemporal latent matching}, that enables SAM to segment changes in bitemporal remote sensing images while inheriting these important properties of SAM (i.e., promptability, zero-shot generalization).
This is achieved by leveraging intra-image and inter-image semantic similarities that we empirically discovered in the latent space of SAM  when applying SAM's encoder on unseen multi-temporal remote sensing images.
The resulting models, \textit{AnyChange}, are capable of segmenting any semantic change. 

AnyChange can yield class-agnostic change masks, however, in some real-world application scenarios, e.g., disaster damage assessment, there is a need for object-centric changes, e.g. to detect how many buildings are destroyed.
To enable this capability we propose a point query mechanism for AnyChange, leveraging SAM's point prompt mechanism and our bitemporal latent matching for filtering desired object changes.
The user only needs a single click on a desired object, AnyChange with the point query can yield change masks centered on this object's semantics, i.e., from this object class to others and vice versa, thus achieving object-centric change detection.

\begin{figure*}[ht]
  \centering
   \includegraphics[width=.9\linewidth]{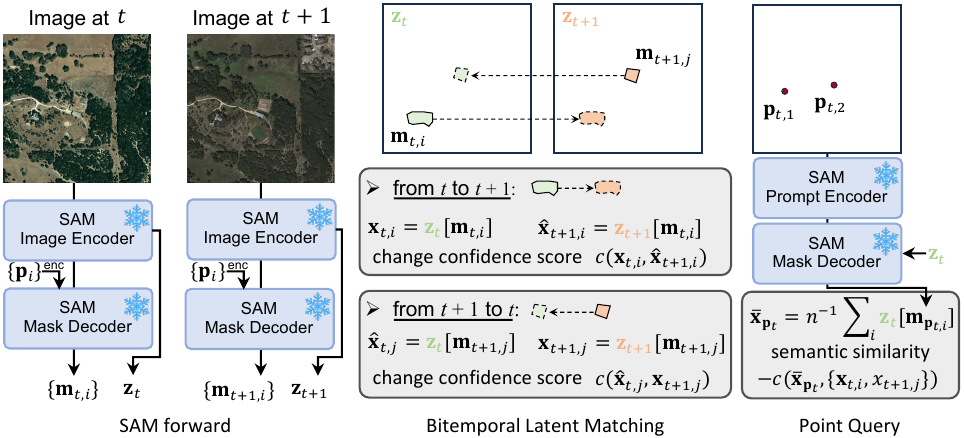}
   \vspace{-0em}
   \caption{\textbf{Segment Any Change Models, AnyChange}. 
   SAM forward: given grid points $\{\mathbf{p}_i\}$ as prompts and input images, SAM produces object masks $\{\mathbf{m}_{t,i}\}$ and image embedding $\mathbf{z_t}$ on the image at time $t$.
   \textbf{Bitemporal Latent Matching} does a bidirectional matching to compute the change confidence score for each change proposal, and then top-k sorting or thresholding is applied for zero-shot change proposal and detection.
   \textbf{Point Query} allows users to click some points (the case of two points in this subfigure) with the same category to filter class-agnostic change masks via semantic similarity for object-centric change detection.
   }
   \label{fig:anychange}
   \vspace{-1em}
\end{figure*}

We demonstrate the zero-shot prediction capabilities of AnyChange on several change detection datasets, including LEVIR-CD \citep{levircd}, S2Looking \citep{shen2021s2looking}, xView2 \citep{gupta2019xbd}, and SECOND \citep{second_dataset}.
Due to the absence of published algorithms for zero-shot change detection, we also build baselines from the perspectives of zero-shot change proposal, zero-shot object-centric change detection, and unsupervised change detection.
AnyChange outperforms other zero-shot baselines implemented by DINOv2 \citep{dinov2} and SAM with different matching methods in terms of zero-shot change proposal and detection.
From the unsupervised change detection perspective, AnyChange beats the previous state-of-the-art model, I3PE \citep{I3PE}, setting a new record of 48.2\% F$_1$ on SECOND.
We show some qualitative results in Fig.~\ref{fig:teaser}, demonstrating the zero-shot prediction capabilities of AnyChange on a wide range of application scenarios (i.e., urbanization, disaster damage assessment, de-agriculturalization, deforestation, and natural resource monitoring).
The contributions of this paper are summarized as follows:
\vspace{-2mm}
\begin{itemize}[leftmargin=*]
\item \textbf{AnyChange}, the first zero-shot change detection model, enables us to obtain both instance-level and pixel-level change masks either in a fully automatic mode or interactively with simple clicks.
\vspace{-1mm}
\item \textbf{Bitemporal Latent Matching}, a training-free adaptation method, empowers SAM with zero-shot change detection by leveraging intra-image and inter-image semantic similarities of images in SAM's latent space. 
\vspace{-1mm}  
\item \textbf{Zero-Shot Change Detection} is explored for the first time. We demonstrate the effectiveness of AnyChange from four perspectives, i.e., zero-shot change proposal, zero-shot object-centric change detection, unsupervised change detection, and label-efficient supervised change detection, achieving better results over strong baselines or previous SOTA methods.
\end{itemize}

\section{Related Work}
\label{related_work}

\textbf{Segment Anything Model} \citep{sam} is the first foundation model for promptable image segmentation, possessing strong zero-shot generalization on unseen object types, data distributions, and tasks.
The training objective of SAM is to minimize a class-agnostic segmentation loss given a series of geometric prompts.
Based on compositions of geometric prompts, i.e., prompt engineering, SAM can generalize to unseen single-image tasks in a zero-shot way, including edge detection, object proposal, and instance segmentation.
Our work extends SAM with zero-shot change detection for bitemporal remote sensing images via a training-free adaptation method, extending the use of SAM beyond single-image tasks.

\textbf{Segment Anything Model for Change Detection.}
SAM has been used for change detection via a ``parameter-efficient fine-tuning'' (PEFT) paradigm \citep{peft}, such as SAM-CD \citep{ding2023adapting} that used Fast-SAM \citep{fastsam} as a frozen visual encoder and fine-tuned adapter networks and the change decoder on change detection datasets in a fully supervised way.
This model does not inherit the most two important properties of SAM, i.e., promptability and zero-shot generalization. 
Fine-tuning in a promptable way with large-scale training change data may achieve these two properties, however, collecting large-scale bitemporal image pairs with class-agnostic change annotations is non-trivial \citep{tewkesbury2015critical,changen}, thus no such method exists in the current literature.
Our work introduces a new and economic adaptation method for SAM, i.e., training-free adaptation,  guaranteeing these two properties with zero additional cost, making zero-shot change detection feasible for the first time.

\textbf{Unsupervised Change Detection.}
The most similar task to zero-shot change detection is unsupervised change detection \citep{coppin1996digital}, however zero-shot change detection is a more challenging task.
They both require models to find class-agnostic change regions, and the main difference is that zero-shot change detection also requires models to generalize to unseen data distributions.
From early model-free change vector analysis (CVA) \citep{bruzzone2000automatic, bovolo2006theoretical} to advanced deep CVA \citep{DCVA} and I3PE \citep{I3PE}, unsupervised change detection methods have undergone a revolution enabled by deep visual representation learning.
These model-based unsupervised change detection methods need to re-train their models on new data distributions.
Our proposed model is training-free and can achieve comparable or even better performance for unsupervised change detection.

\section{Segment Any Change Models}
\label{sec:method}

This paper introduces \textit{Segment Any Change} to resolve the long-standing open problem of zero-shot change detection (see Sec.~\ref{sec:sacm:background}).
As illustrated in Fig.~\ref{fig:anychange}, we propose a new type of change detection models that support zero-shot prediction capabilities and generalization on unseen change types and data distributions, allowing two output structures (instance-level and pixel-level), and three application modes (fully automatic, semi-automatic with a custom threshold, and interactive with simple clicks).
AnyChange achieves the above capabilities building on SAM in a training-free way.

\vspace{-2mm}
\subsection{Background}
\label{sec:sacm:background}

\vspace{-1mm}
\textbf{Preliminary: Segment Anything Model} (SAM) is a promptable image segmentation model with an image encoder of ViT \citep{vit}, a prompt encoder, and a mask decoder based on transformer decoders.
As Fig.~\ref{fig:anychange}(a) shows, given as input a single image at time $t$, the image embedding $\mathbf{z}_t$ is extracted from its image encoder.
Object masks $\{\mathbf{m}_{t,i}\}$ can be obtained by its mask decoder given $\mathbf{z}_t$ and dense point prompts obtained by feeding grid points $\{\mathbf{p}_i\}$ into its prompt encoder.

\vspace{-1mm}
\textbf{Problem Formulation: Zero-Shot Change Detection} is formulated at the pixel and instance levels.
Our definition of ``zero-shot'' is similar to that of SAM, i.e., the model without training on change detection tasks can transfer zero-shot to change detection tasks and new image distributions.
$\mathcal{U}=\{0, 1\}$ denotes a set of classes of non-change and change.
A pixel position or instance area belongs to the change class if the corresponding semantic categories at different times are different.
This means that the model should generalize to unseen change types, even though all change types are merged into a class of 1.
The input is a bitemporal image pair $\I_t, \I_{t+1} \in \mathbb{R}^{h\times w\times *}$, where the image size is $(h, w)$ and $*$ denotes the channel dimensionality of each image.
For pixel level, the model is expected to yield a pixel-level change mask $\mathbf{C}\in\mathcal{U}^{h\times w}$.
For instance level, the model is expected to yield an arbitrary-sized set of change masks ${\{\mathbf{m}_i}\}$, where each instance $\mathbf{m}_i$ is a polygon.

\subsection{Exploring the Latent Space of SAM}
Motivated by the experiments in \citet{sam} on probing the latent space of SAM, we known there are potential semantic similarities between mask embeddings in the same natural image.
We further explore the latent space for satellite imagery from both intra-image and inter-image perspectives, thus answering the following two questions:

\begin{figure}[htbp]
\centering
\begin{minipage}[t]{0.4\linewidth}
\centering
\includegraphics[width=1.\linewidth]{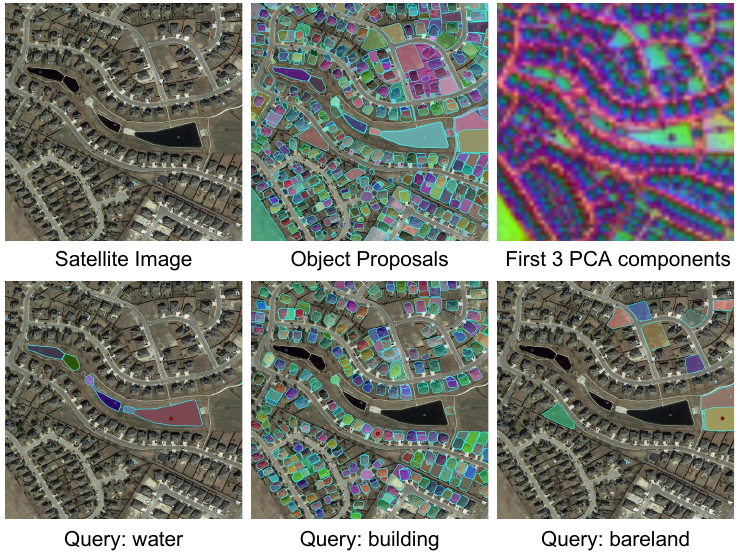}
\subcaption{intra-image semantic similarity}
\end{minipage}
\quad
\begin{minipage}[t]{0.513\linewidth}
\centering
\includegraphics[width=1.\linewidth]{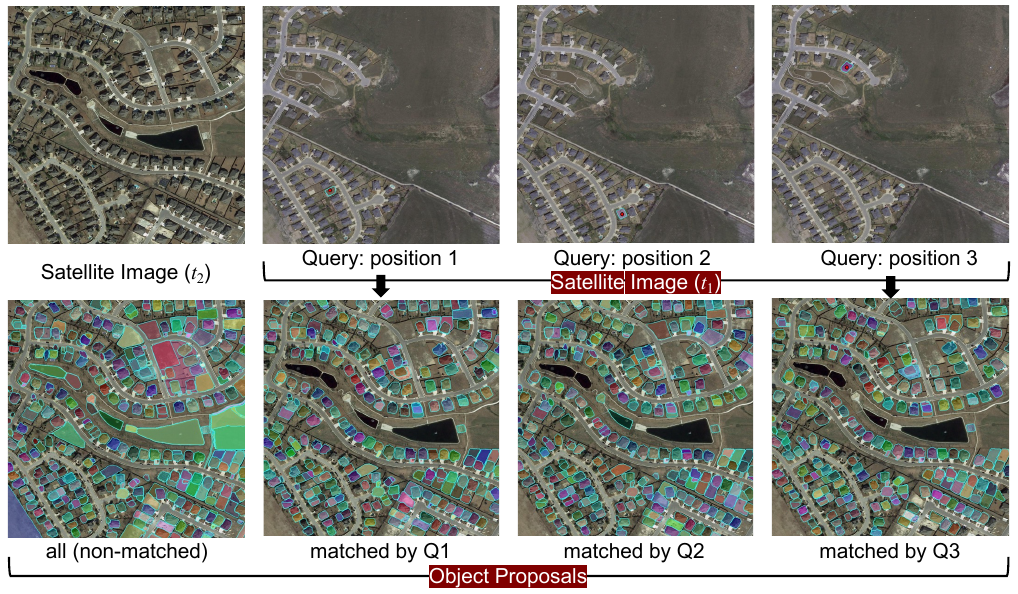}
\subcaption{inter-image semantic similarity}
\end{minipage}
\caption{Empirical evidence on semantic similarity on (a) the same satellite image and (b) the satellite images at different times.
\textbf{(a)} The visualization of the first three PCA components indicates the objects with the same category are well matched with each other in SAM's latent space; Three \textcolor{red}{red} point queries confirm the existence of semantic similarity on the same satellite image.
\textbf{(b)} The object proposals (indicated by \textcolor{red}{red} points) from the satellite image at $t_1$ as queries are used to match all proposals from the satellite image at $t_2$.
(best viewed with zoom, especially for the point query)}
\label{fig:latent}
\vspace{-7mm}
\end{figure}

\textit{Q1: Do semantic similarities exist on the same satellite image?}
Empirically, we find they do.
We show this semantic similarity in two ways, i.e., visualizing the first components of principal components analysis (PCA) \citep{dinov2} and probing the latent space \citep{sam}, as Fig.~\ref{fig:latent} (a) shows.
Observing the first three PCA components, geospatial objects with the same category have a similar appearance in this low-dimensional subspace.
This suggests that this satellite image embedding from SAM encodes the semantics of geospatial objects reasonably well.
Furthermore, we do the same latent space probing experiment as \citet{sam} did, but on satellite images, and present the results in Fig.~\ref{fig:latent} (a) (bottom).
We compute the mask embedding of object proposals and manually select three proposals with different categories (water, building, and bareland) as queries.
By matching the query embedding with other mask embeddings, we obtain the most similar object proposals with the query proposal.
We find that the most similar object proposals mostly belong to the same category as the query.

\textit{Q2: Do semantic similarities exist on satellite images of the same location collected at different times?}
Empirically, we find they do.
To verify this, we introduce a new satellite image from the same geographic area but at a different time $t_1$.
The above satellite image is captured at time $t_2$.
Different from the above latent space probing experiment, we use three object proposals with different spatial positions from the image at $t_1$, as queries.
These three queries have the same category, i.e., building.
By matching with all proposals from the image at $t_2$, we obtain three basically consistent results (F$_1$ of 68.1\%$\pm$0.67\% and recall of 96.2\%$\pm$0.66\%), as shown in Fig.~\ref{fig:latent} (b).
This suggests that this semantic similarity exists on the satellite images at different times, even though the images have different imaging conditions because they were taken at different times.

From the above empirical study, we find that there are intra-image and inter-image semantic similarities in the latent space of SAM for unseen satellite images.
These two properties are the foundation of our training-free adaptation method.

\vspace{-3mm}
\subsection{Bitemporal Latent Matching}\vspace{-1mm}
Based on our findings, we propose a training-free adaptation method, namely \emph{bitemporal latent matching}, which bridges the gap between SAM and remote sensing change detection without requiring for training or architecture modifications.

\vspace{-1mm}
The main idea is to leverage the semantic similarities in latent space to identify  changes in bitemporal satellite images.
Given the image embedding $\mathbf{z}_t$ and object proposals $\mathcal{M}_{t} = \{\mathbf{m}_{t, i}\}_{i \in [1, 2, ..., N_t]}$ generated from SAM on the satellite image at time $t$, each object proposal $\mathbf{m}_{t, i} \in \mathbb{R}^{h\times w}$ is a binary mask.
We can compute the mask embedding $\mathbf{x}_{t, i} = \mathbf{z}_t[\mathbf{m}_{t, i}] \in R^{d_m}$ by averaging the image embedding $\mathbf{z}_t$ over all non-zero positions indicated by the object proposal $\mathbf{m}_{t, i}$.
Next, we introduce a similarity metric to measure the semantic similarity.
To do this, we need to consider the statistical properties of the image embeddings from SAM.
 In particular, SAM's image encoder uses layer normalization \citep{LN}.
This means that the mask embedding has zero mean and unit variance if we drop the affine transformation in the last layer normalization, i.e., the variance $D(\mathbf{x}_{t, i}) = d_m^{-1}\sum_j(\mathbf{x}_{t, i}[j])^2 = 1$, thus we have the mask embedding's $\ell_2$ norm $\left\lVert\mathbf{x}_{t, i}\right\rVert_2 = \sqrt{d_m}$, which is a constant since $d_m$ is the channel dimensionality.
Given this, cosine similarity is a suitable choice to measure similarity between two mask embeddings since they are on a hypersphere with a radius of $\sqrt{d_m}$, and differences are encoded by their directions.
Therefore, we propose to use negative cosine similarity as the change confidence score $c(\mathbf{x}_i, \mathbf{x}_j)$ for mask embeddings $\mathbf{x}_i$ and $\mathbf{x}_j$:
\begin{equation}
  c(\mathbf{x}_i, \mathbf{x}_j) = -\frac{\mathbf{x}_{i}\cdot\mathbf{x}_{j}}{d_m}
\end{equation}

The next question is which two mask embeddings to use to compute the change confidence score.
The real-world change is defined at the same geographic location from time $t$ to $t+1$.
This means that it is comparable only if two mask embeddings cover the approximate same geographic region.
Therefore, we additionally compute the mask embedding $\hat{\mathbf{x}}_{t+1, i} = \mathbf{z}_{t+1}[\mathbf{m}_{t, i}] \in \mathbb{R}^{d_m}$ on the image embedding $\mathbf{z}_{t+1}$ using the same object proposal $\mathbf{m}_{t, i}$.
We then compute the change confidence score $c(\mathbf{x}_{t, i}, \hat{\mathbf{x}}_{t+1, i})$ for the change at $\mathbf{m}_{t, i}$ from $t$ to $t+1$.

Since we need to guarantee the temporal symmetry \citep{changestar, zheng2022changemask} of class-agnostic change, we propose to match the object proposals bidirectionally.
To this end, the change confidence score $c(\mathbf{x}_{t+1, i}, \hat{\mathbf{x}}_{t, i})$ for the change at $\mathbf{m}_{t+1, i}$ from $t+1$ to $t$ is also computed, where $\hat{\mathbf{x}}_{t, i} = \mathbf{z}_{t}[\mathbf{m}_{t+1, i}] \in \mathbb{R}^{d_m}$ is computed on the image embedding $\mathbf{z}_{t}$ with the same object proposal $\mathbf{m}_{t+1, i}$.
Afterwards, we can match object proposals $\mathcal{M}_t$ and $\mathcal{M}_{t+1}$ bidirectionally, and $(N_t + N_{t+1})$ change proposals with their confidence score are obtained in total.
We propose to finally obtain change detection predictions by sorting by confidence scores and selecting the top-k elements or by angle thresholding.
The pseudo-code of Bitemporal Latent Matching is in Appendix~\ref{supp:pseudocode}.

\vspace{-2mm}
\subsection{Point Query Mechanism}\vspace{-1mm}
To empower AnyChange with interactive change detection with semantics, we combine our bitemporal latent matching with the point prompt mechanism of SAM, thus yielding the point query mechanism.
Given as input a set of single-temporal point $\{\mathbf{p}_{t,i}\} = {(x_{t,i}, y_{t,i})}$ with the same category, $t$ denote that this point belongs to the image at time $t$, and $(x,y)$ indicates the spatial coordinate of image domain.
The object proposals $\{\mathbf{m}_{\mathbf{p}_{t,i}}\}$ can be obtained via SAM's point prompts.
Following our bitemporal latent matching, we then compute their average mask embedding $\bar{\mathbf{x}}_{\mathbf{p}_t} = n^{-1}\sum_1^n\mathbf{x}_{\mathbf{p}_{t, i}}$ and match it with all proposals $\{\mathcal{M}_{t}, \mathcal{M}_{t+1}\}$ via cosine similarity.
In this way, the object-centric change detection results can be obtained via a custom angle threshold.

\vspace{-4mm}
\section{Experiments}
\label{exp}\vspace{-2mm}
In this section, we demonstrate the two most basic applications of AnyChange, i.e., (i) zero-shot change proposal and detection and (ii) change data engine.
We conduct experiments from these two perspectives to evaluate our method.

\vspace{-2mm}
\subsection{Zero-Shot Object Change Proposals}
\textbf{Datasets:}
We use four commonly used change detection datasets to evaluate  AnyChange.
The first three, i.e., LEVIR-CD \citep{levircd}, S2Looking \citep{shen2021s2looking}, and xView2 \citep{gupta2019xbd}, are building-centric change detection datasets.
SECOND \citep{second_dataset} is a multi-class (up to 36 change types) urban change detection dataset with full annotation.
For zero-shot object proposal evaluation, we convert their labels into binary if the dataset has multi-class change labels.
\vspace{-2mm}

\textbf{Metrics:}
Conventional change detection mainly focuses on pixel-level evaluation using F$_1$, precision, and recall.
More and more real-world applications have started to focus on instance-level change, also called ``change parcel''.
Based on this requirement and AnyChange's capability of predicting instance change, we adapt the evaluation protocol of the zero-shot object proposal \citep{sam} for the zero-shot change proposal since they have the same output structure.
The metric is mask AR@1000 \citep{mscoco}.
Note that change proposals are class-agnostic, therefore, for the first three building-centric datasets, we cannot obtain accurate F$_1$ and precision due to incomplete ``any change'' annotations.
These two metrics are only used for reference and to see whether the model is close to the naive baseline (predict all as “change” class, and vice versa). 
Here we mainly focus on recall for both pixel- and instance-levels.
\vspace{-2mm}

\begin{table*}[ht]
    \caption{\textbf{Zero-shot Object Change Proposals}.
    The metrics include \textcolor{pix}{pixel-based} F$_1$, Precision (Prec.), and Recall (Rec.) and \textcolor{ins}{instance-based} mask AR@1000.
    Note that the metric names rendered with \textcolor{inacc}{gray} represent inaccurate estimations due to the absence of ground truth of ``any change'', but reflect whether their predictions approximate the naive baseline (predict all as ``change'' class).
    \label{tab:benchmark:zsocp}}
    \centering
    \tablestyle{2pt}{1.2}
    \resizebox{\linewidth}{!}{
    \begin{tabular}{lc>{\color{gray}}c>{\color{gray}}ccc|>{\color{gray}}c>{\color{gray}}ccc|>{\color{gray}}c>{\color{gray}}ccc|cccc}
      \toprule
                & &\multicolumn{4}{c}{\bf LEVIR-CD} & \multicolumn{4}{c}{\bf S2Looking {\rm\scriptsize(binary)}} & \multicolumn{4}{c}{\bf xView2 {\rm\scriptsize(binary)}} & \multicolumn{4}{c}{\bf SECOND {\rm\scriptsize(binary)}}   \\
      Method       &  Backbone  & \cellcolor{pix} F$_1$ & \cellcolor{pix}Prec. & \cellcolor{pix}Rec. & \cellcolor{ins}{\scriptsize mask AR}$\uparrow$ & \cellcolor{pix} F$_1$ & \cellcolor{pix}Prec. & \cellcolor{pix}Rec. & \cellcolor{ins}{\scriptsize mask AR}$\uparrow$ & \cellcolor{pix} F$_1$ & \cellcolor{pix}Prec. & \cellcolor{pix}Rec. & \cellcolor{ins}{\scriptsize mask AR}$\uparrow$ & \cellcolor{pix} F$_1\uparrow$ & \cellcolor{pix}Prec. & \cellcolor{pix}Rec. & \cellcolor{ins}{\scriptsize mask AR}$\uparrow$            \\
      \hline
      \pixb \textit{pixel level} & & & & &  & & & & &  & & & & &  & & \\
      CVA& - &12.2&7.5&32.6& - & 5.8 & 3.1 & 44.3 & - & 7.6 & 4.3 & 33.3 & - & 30.2 & 26.5 & 35.2 & - \\
      DINOv2+CVA& ViT-G/14& 17.3 & 9.5 & 96.6 & - & 4.3 & 2.2 & 92.9 & - & 5.9 & 3.1 & 62.0 & - & 41.4 & 26.9 & 89.4 & - \\
      \multicolumn{3}{l}{\insb \textit{pixel and instance level}} & & &  & & & & &  & & & & &  & &\\
      SAM+Mask Match  &  ViT-B  &  12.2 &  8.7  &  20.2  &  6.8  &  4.7  &  2.6 & 28.6 &  15.1  &  8.6  & 8.7 &  23.5  &  10.2 & 23.5 & 30.3 & 19.2 & 7.2 \\
      SAM+CVA Match   &  ViT-B  &  12.7 &  7.5  &  41.9  &  9.0  &  3.7  & 1.9  & 78.3 &  31.5  &  3.0  & 1.6 &  29.3  &  12.2 & 34.1 & 23.9 & 59.5 & 18.6 \\
      {\bf AnyChange} &  ViT-B  &  23.4 &  13.7 &  83.0  & \best{32.6} &  7.4  &  3.9 & 94.0 & \best{48.3} &  13.4 & 7.6 &  59.3  & \best{27.8} & \best{44.6} & 30.5 & 83.2 & \best{27.0} \\
      \gr\gc{AnyChange (Oracle)} & ViT-B & \bt{73.3}  & \bt{65.2}  & 83.6  & 37.3 & \bt{60.3} & \bt{53.4} & 69.1 & 29.8 & \bt{49.7} & \bt{38.7} & 69.5  & 31.6 & 69.5 & 68.2 & 70.8  &  16.8  \\
      \hline
      SAM+Mask Match  &  ViT-L  &  16.8 &  11.9 &  28.8  &  13.4 &  4.0  &  2.3  & 13.6 &  8.5   &  6.8  & 4.3 &  15.7  &  7.1  & 16.0 & 29.0 & 11.0 & 4.2    \\
      SAM+CVA Match   &  ViT-L  &  13.2 &  7.2  &  88.2  &  29.7 &  3.0  & 1.5   & 95.5 &  40.3  &  2.6  & 1.3 &  26.6  &  11.6 & 35.3 & 22.0 & 90.1 & 25.5   \\
      {\bf AnyChange} &  ViT-L  &  21.9 &  12.5 &  87.1  & \best{39.5} &  6.5  &  3.3  & 93.0 & \best{50.1}  &  9.8  & 5.3 &  66.1  & \best{30.5} & \textbf{42.3} & 27.9 & 87.4 & \best{28.6}   \\
      \gr\gc{AnyChange (Oracle)} & ViT-L & \bt{75.3}  & \bt{65.4}  & 88.7  & 44.8 & \bt{62.2} & \bt{57.6} & 67.6 & 30.6 & \bt{51.4} & \bt{37.5} & 81.7 & 38.8 & 69.1 &63.0 & 76.5 & 19.9 \\
      \hline
      SAM+Mask Match  &  ViT-H  &  17.8 &  12.6 &  30.2  &  16.1 &  4.1  &  2.5  & 13.1 &  8.6   &  5.5  & 3.5  &  13.3  &  6.3  & 14.2 & 28.8 & 9.5  & 3.7   \\
      SAM+CVA Match   &  ViT-H  &  13.2 &  7.1  &  92.3  &  36.8 &  2.9  &  1.5  & 96.3 &  41.2  &  3.1  & 1.6  &  34.1  &  15.7 & 35.6 & 22.1 & 91.1 & 25.7  \\
      {\bf AnyChange} &  ViT-H  & 23.0  &  13.3 &  85.0  & \best{43.4} & 6.4   &  3.3  &  93.2 & \best{50.4}  & 9.4  & 5.1 & 62.2  &  \best{29.3} & \best{41.8} & 27.4 & 88.7 & \best{29.0}  \\
      \gr\gc{AnyChange (Oracle)}& ViT-H & \bt{76.3} & \bt{64.7} & 93.1 & 51.4 & \bt{61.0} & \bt{53.2} & 71.3 & 31.6 & \bt{50.8} & \bt{36.4} & 84.2 & 40.4 & 69.5 & 63.7 & 76.5 & 19.3 \\
      \bottomrule
    \end{tabular}
    }
    \vspace{-1em}
\end{table*}

\textbf{Baselines:}
AnyChange is based on SAM, however, there is no SAM or other VFM\footnote{Visual foundation model, like SAM, CLIP, and DINOv2, etc.}-based zero-shot change detection model that can be used for comparison in the current literature.
For a fair comparison, we build three strong baselines based on DINOv2 \citep{dinov2} (a state-of-the-art VFM) or SAM.
The simple baseline is CVA \citep{bruzzone2000automatic}, a model-free unsupervised change detection method based on $\ell_2$ norm as dissimilarity and thresholding.
We build ``DINOv2+CVA'', an improved version with DINOv2 using an idea similar to DCVA \citep{DCVA}.
We build ``SAM+Mask Match'', which follows the macro framework of AnyChange and replaces the latent match with the mask match that adopts the IoU of masks as the similarity.
``SAM+CVA Match'' follows the same idea of AnyChange but adopts the negative $\ell_2$ norm of feature difference as the similarity to compute pixel-level change map via SAM feature-based CVA first.
The instance-level voting is then adopted to obtain change proposals.
We also build each ``Oracle'' version of AnyChange as an upper bound, where we fine-tune SAM with LoRA ($r=32$) \citep{hu2022lora} and train a change confidence score network on each dataset.
More implementation details can be seen in Appendix~\ref{supp:baseimpl}.
\vspace{-2mm}

\textbf{Results:}
We compare the change recall of AnyChange with other zero-shot baselines in Table~\ref{tab:benchmark:zsocp}.
For \textit{pixel-level change recall}, AnyChange achieves better recalls than the other two SAM baselines, especially when using a small ViT-B.
This is because bitemporal latent matching better measures semantic similarity, i.e., using the angle between two embeddings.
We can observe that the gap between AnyChange and SAM baselines gradually reduces as the backbone becomes larger since visual representation capabilities generally become stronger.
AnyChange still has better average performance on four datasets, although a stronger representation can close the performance gap to some extent.
This highlights the importance of finding the essential semantic difference.
Besides, our AnyChange with ViT-H (636M parameters) achieves comparable recalls to CVA with DINOv2 (ViT-G, 1,100M parameters) on four datasets in fewer parameters.
On the SECOND dataset, all variants of AnyChange achieve better zero-shot change detection performance than the strong baseline, DINOv2+CVA, and the margin is up to 3.2\% F$_1$.
For \textit{instance-level change recall}, AnyChange outperforms the other two SAM baselines by a significant margin.
This further confirms the effectiveness and superiority of bitemporal latent matching.
We observe that the Oracles obtained via supervised learning have superior precision since they learn semantics and dataset bias explicitly, however, this comes with the cost of recall on pixel or instance levels.
\vspace{-1mm}

\textbf{Ablation: Matching Strategy.}
``Mask Match'' performs geometry-based matching, while ``CVA Match'' and bitemporal latent matching perform latent-based matching.
In Table~\ref{tab:benchmark:zsocp}, we see that latent-based matching is much more promising than geometry-based matching from multiple perspectives of pixel-level and instance-level change recall and zero-shot change detection performance.
Compared with ``CVA Match'', AnyChange outperforms ``SAM+CVA Match'' on instance-level object change proposals by large margins and has comparable recalls on pixel-level change recalls.
In principle, the similarity of ``CVA Match'' is linearly related to our bitemporal latent matching since the magnitude of the embeddings is a constant.
Therefore, the performance difference lies in the computational unit (pixel or instance). 
``CVA Match'' is pixel-wise, while bitemporal latent matching is instance-wise.
This suggests that it is more robust to compute mask embedding averaged over all involved pixel embeddings for matching.

\textbf{Ablation: Matching Direction.}
As presented in Table~\ref{tab:ab:md}, we can find that the performance of single-directional matching is sensitive to temporal order, e.g., mask AR of two single-directional matching on LEVIR-CD are 1.3\% and 35.9\%, respectively. 
This is because the class-agnostic change is naturally temporal symmetric which is exactly the motivation of our bidirectional design. 
This result also confirms generally higher and more robust zero-shot change proposal capability.

\textbf{Ablation: Robustness to radiation variation.}
We used ViT-B as the backbone for fast experiments. 
The results are presented in Table~\ref{tab:ab:rrv}. 
The performance jitter of mask AR is less than 2\% (-1.9\%,+0.1\%, -1.9\%, -1.4\%) on these four datasets. 
We believe this sensitivity to radiation variation is acceptable for most applications.

\begin{table*}[t]
    \caption{\textbf{Ablation: Matching Direction}.
    The backbones are ViT-B.
    Single-directional matching is sensitive to temporal order, while bidirectional matching is more stable due to its guaranteed temporal symmetry.
    \label{tab:ab:md}}
    \centering
    \tablestyle{2pt}{1.2}
    \resizebox{\linewidth}{!}{
    \begin{tabular}{l>{\color{gray}}c>{\color{gray}}ccc|>{\color{gray}}c>{\color{gray}}ccc|>{\color{gray}}c>{\color{gray}}ccc|cccc}
      \toprule
     &\multicolumn{4}{c}{\bf LEVIR-CD} & \multicolumn{4}{c}{\bf S2Looking {\rm\scriptsize(binary)}} & \multicolumn{4}{c}{\bf xView2 {\rm\scriptsize(binary)}} & \multicolumn{4}{c}{\bf SECOND {\rm\scriptsize(binary)}}   \\
      Direction     & \cellcolor{pix} F$_1$ & \cellcolor{pix}Prec. & \cellcolor{pix}Rec. & \cellcolor{ins}{\scriptsize mask AR}$\uparrow$ & \cellcolor{pix} F$_1$ & \cellcolor{pix}Prec. & \cellcolor{pix}Rec. & \cellcolor{ins}{\scriptsize mask AR}$\uparrow$ & \cellcolor{pix} F$_1$ & \cellcolor{pix}Prec. & \cellcolor{pix}Rec. & \cellcolor{ins}{\scriptsize mask AR}$\uparrow$ & \cellcolor{pix} F$_1\uparrow$ & \cellcolor{pix}Prec. & \cellcolor{pix}Rec. & \cellcolor{ins}{\scriptsize mask AR}$\uparrow$            \\
      \hline
      bidirectional &  23.4 &  13.7 &  83.0  & 32.6 &  7.4  &  3.9 & 94.0 & 48.3 &  13.4 & 7.6 &  59.3  & 27.8 & 44.6 & 30.5 & 83.2 & 27.0 \\
      from $t$ to $t+1$ & 17.7 & 10.1 & 72.8 & 1.3 & 9.0 & 4.8 & 85.6 & 32.1 & 15.3 & 9.0 & 49.8 & 27.3	& 41.2 & 31.2 & 60.6 & 14.8 \\
      from $t+1$ to $t$ & 23.6 & 13.6& 88.7 & 35.9& 8.1 & 4.3 & 79.5 & 19.4& 12.3 & 7.6 & 32.7 & 6.7 & 46.1 & 34.1 & 71.3 & 14.9 \\
      \bottomrule
    \end{tabular}
    }
    \vspace{-1em}
\end{table*}

\begin{table*}[t]
    \caption{\textbf{Ablation: Robustness to radiation variation}.
    The backbones are ViT-B.
    Radiation variation was simulated by applying random color jitter independently to the pre- and post-event images.
    \label{tab:ab:rrv}}
    \centering
    \tablestyle{2pt}{1.2}
    \resizebox{\linewidth}{!}{
    \begin{tabular}{l>{\color{gray}}c>{\color{gray}}ccc|>{\color{gray}}c>{\color{gray}}ccc|>{\color{gray}}c>{\color{gray}}ccc|cccc}
      \toprule
     &\multicolumn{4}{c}{\bf LEVIR-CD} & \multicolumn{4}{c}{\bf S2Looking {\rm\scriptsize(binary)}} & \multicolumn{4}{c}{\bf xView2 {\rm\scriptsize(binary)}} & \multicolumn{4}{c}{\bf SECOND {\rm\scriptsize(binary)}}   \\
      Condition     & \cellcolor{pix} F$_1$ & \cellcolor{pix}Prec. & \cellcolor{pix}Rec. & \cellcolor{ins}{\scriptsize mask AR}$\uparrow$ & \cellcolor{pix} F$_1$ & \cellcolor{pix}Prec. & \cellcolor{pix}Rec. & \cellcolor{ins}{\scriptsize mask AR}$\uparrow$ & \cellcolor{pix} F$_1$ & \cellcolor{pix}Prec. & \cellcolor{pix}Rec. & \cellcolor{ins}{\scriptsize mask AR}$\uparrow$ & \cellcolor{pix} F$_1\uparrow$ & \cellcolor{pix}Prec. & \cellcolor{pix}Rec. & \cellcolor{ins}{\scriptsize mask AR}$\uparrow$ \\
      \hline
      baseline &  23.4 &  13.7 &  83.0  & 32.6 &  7.4  &  3.9 & 94.0 & 48.3 &  13.4 & 7.6 &  59.3  & 27.8 & 44.6 & 30.5 & 83.2 & 27.0 \\
      w/ color jitter & 22.6 & 13.1 & 84.2 & 30.7&7.4 & 3.8 & 94.1 & 48.4& 13.5 & 7.7 & 54.7 & 25.9&42.4 & 28.7 & 81.7 & 26.4 \\
      \bottomrule
    \end{tabular}
    }
    \vspace{-1em}
\end{table*}

\vspace{-2mm}
\subsection{Zero-shot Object-centric Change Detection}\vspace{-1mm}
Zero-shot object change proposal yields class-agnostic change masks.
The point query mechanism can convert the class-agnostic change into object-centric change via simple clicks of the user on a single-temporal image, thus providing an interactive mode for AnyChange.
This step is typically easy for humans. 
Here we evaluate the point query on three building-centric change datasets.
\vspace{-1mm}

\begin{figure}[t]
\centering
\includegraphics[width=0.9\linewidth]{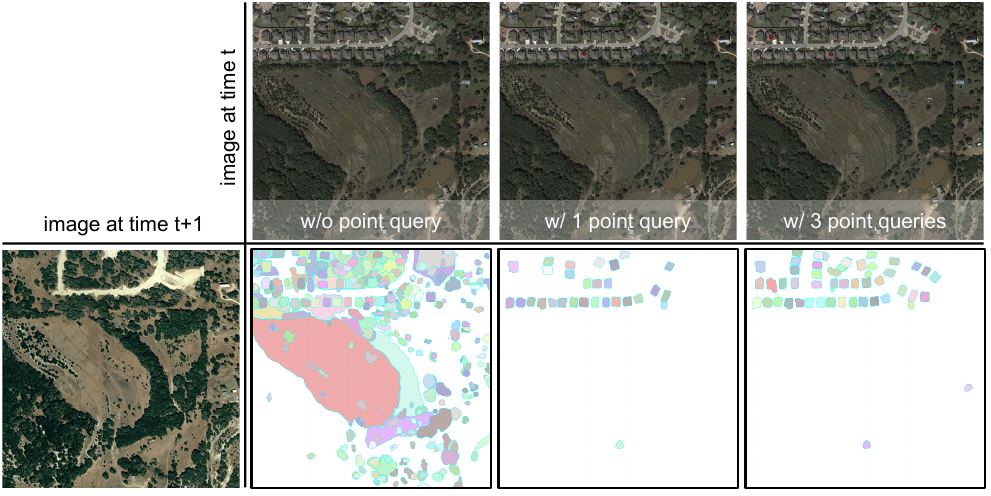}
\caption{Examples of \textbf{Point Query Mechanism}.
The effects of w/o point query, one-point query, and three-point queries are shown in sequence from left to right.
(best viewed digitally with zoom, especially for the \textcolor{red}{red} points)
}
\label{fig:pq}
\vspace{-4mm}
\end{figure}

\textbf{Results:}
We demonstrate the effect of the point query in Fig.~\ref{fig:pq}.
Without the point query, AnyChange yields class-agnostic change masks, including building changes, vegetation changes, etc. 
With a single-point query on a building, we can observe that change masks unrelated to the building are filtered out.
Further clicking two more buildings to improve the stability of mask embedding, we find the building changes previously missed are successfully recovered.
Table~\ref{tab:benchmark:single_point} quantitatively reflects this mechanism.
With a single-point query, the zero-shot performances on three datasets significantly gain $\sim 15\%$ F$_1$ score.
This improvement hurts recall as a cost, however it achieves a better trade-off between precision and recall.
\begin{wraptable}{r}{0.68\textwidth}
\vspace{-3.5mm}
\caption{\textbf{Zero-shot Object-centric Change Detection.}
All results of introducing semantics from the point query are accurate estimations since the detected changes are object-centric.
\label{tab:benchmark:single_point}}
\vspace{-2mm}
\centering
\small
\tablestyle{2pt}{1.2}
\begin{tabular}{lccc|ccc|ccc}
\toprule
          & \multicolumn{3}{c}{\bf LEVIR-CD} & \multicolumn{3}{c}{\bf S2Looking}{\rm\scriptsize(binary)}  & \multicolumn{3}{c}{\bf xView2{\rm\scriptsize(binary)}}  \\
Method    & F$_1$ & Prec. & Rec. & F$_1$ & Prec. & Rec. & F$_1$ & Prec. & Rec. \\
\hline
AnyChange-H             & 23.0      & 13.3 & 85.0 & 6.4  & 3.3  & 93.2 & 9.4   &  5.1 &  62.2  \\
+ Point Query           &           &      &      &      &      &      &       &      &        \\
1 point query          & 38.6      & 30.9 & 51.4 & 21.8 & 16.9 & 34.1 & 25.1  & 24.8 &  25.4  \\
                        & \up{15.6} &      &      & \up{15.4} & &      & \up{15.7} &  &        \\
3 point queries         & 42.2      & 28.5 & 81.1 & 24.0 & 14.5 & 68.5 & 28.1  & 20.0 &  47.2  \\
                        & \up{19.2} &      &      & \up{17.6} & &      & \up{18.7}    &        \\
\bottomrule
\end{tabular}
\end{wraptable}

\vspace{-5mm}
After increasing to three-point queries, the recalls of the model on three datasets get back to some extent, and the model has comparable precision with the single-point query.
The zero-shot performances on three datasets gain $\sim 3\%$ F$_1$ further.
These results confirm the effectiveness of the point query as a plugin for AnyChange to provide an interactive mode.

\vspace{-2mm}
\subsection{AnyChange as Change Data Engine}\vspace{-1mm}
AnyChange can provide pseudo-labels for unlabeled bitemporal image pairs with zero or few annotation costs.
To evaluate this capability, we conduct experiments on two typical tracks for remote sensing change detection: supervised object change detection and unsupervised change detection.

\textbf{Training recipe:}
On S2Looking, we train the model on its training set with pure pseudo-labels of AnyChange with ViT-H with a single-point query.
All training details follow \citet{changen} except the loss function.
On SECOND, we train the model on its training set with pure pseudo-labels of AnyChange with ViT-H via fully automatic mode, and other details follow \citet{zheng2022changemask}.
For both these two cases, the loss function is BCE loss with a label smoothing of 0.4.
See Appendix~\ref{supp:psetrain} for more details.

\begin{table*}[t]
  \caption{\textbf{Supervised Object Change Detection}. Comparison with the state-of-the-art change detectors on the \textbf{S2Looking} \texttt{test}.
  ``R-18'': ResNet-18.
  The amount of Flops was computed with a float32 tensor of shape [2,512,512,3] as input.
    \label{tab:sota_s2l}}
  \centering
  \small
  \tablestyle{4pt}{1.2}
  \resizebox{\linewidth}{!}{
  \begin{tabular}{l|ccc|ccc|cc}
    \toprule
     Method                             & Backbone    &  Fine-tuning on & \#labeled pixels$\downarrow$  & F$_1\uparrow$         & Prec. & Rec. & \#Params. & \#Flops \\
     \hline
     FC-Siam-Diff \citep{daudt2018fully} & -           & 100\% GT  & $1.2\times10^{10}$ & 13.1          & 83.2  & 15.7 & 1.3M       & 18.7G   \\
     STANet \citep{levircd}              & R-18        & 100\% GT  & $1.2\times10^{10}$ & 45.9          & 38.7  & 56.4 & 16.9M      & 156.7G  \\
     CDNet \citep{chen2021adversarial}   & R-18        & 100\% GT  & $1.2\times10^{10}$ & 60.5          & 67.4  & 54.9 & 14.3M      & -       \\
     BiT \citep{chen2021remote}          & R-18        & 100\% GT  & $1.2\times10^{10}$ & 61.8          & 72.6  & 53.8 & 11.9M      & 34.7G   \\
     ChangeStar (1$\times$96) \citep{changestar}& R-18 & 100\% GT  & $1.2\times10^{10}$ & 66.3          & 70.9  & 62.2 & 16.4M      & 65.3G   \\
    + \texttt{Changen-90k} \citep{changen}   & R-18    & 100\% GT  & $1.2\times10^{10}$ & 67.1          & 70.1  & 64.3 & 16.4M      & 65.3G   \\
    ChangeStar (1$\times$96) \citep{changestar}& MiT-B1& 100\% GT  & $1.2\times10^{10}$ & 64.3          & 69.3  & 59.9  & 18.4M     & 67.3G   \\
    + \texttt{Changen-90k} \citep{changen} & MiT-B1    & 100\% GT  & $1.2\times10^{10}$ & 67.9          & 70.3  & 65.7  & 18.4M     & 67.3G    \\
    \hline
    ChangeStar (1$\times$96)        & R-18            & 1\% GT    & $1.4\times10^{8}$ & 37.2 &  63.1  & 26.3 & 16.4M      & 65.3G      \\
    ChangeStar (1$\times$96)        & R-18            & 0.1\% GT  & $4.7\times10^{6}$ & 9.2  &  10.0  & 8.5  & 16.4M      & 65.3G     \\
    ChangeStar (1$\times$96) (Ours) & R-18            & AnyChange & $3.5\times10^{3}$ & 40.2 &  40.4  & 39.9 & 16.4M      & 65.3G      \\
    \bottomrule
  \end{tabular}}
  \vspace{-5mm}
\end{table*}

\textbf{Supervised Object Change Detection Results.}
Table~\ref{tab:sota_s2l} presents standard benchmark results on the S2Looking dataset, which is the one of most challenging building change detection datasets.
From a data-centric perspective, we use the same architecture with different fine-tuned data.
The model architecture adopts ResNet18-based ChangeStar (1$\times$96) \citep{changestar} due to its simplicity and good performance.
We set the fine-tuning on 100\% ground truth as the upper bound, which achieved 66.3\% F$_1$.
We can observe that fin-tuning on the pseudo-labels of AnyChange with ViT-H yields 40.2\% F$_1$ with 3,500-pixel annotations\footnote{The training set of S2Looking has 3,500 image pairs. We apply AnyChange with a single-point query for each pair to produce pseudo-labels. Therefore, the number of manual annotations is 3,500 pixels.}.
Leveraging AnyChange, we achieve 61\% of the upper bound at a negligible cost ($\sim 10^{-5}$\% full annotations).
We also compare it to the model trained with fewer annotations (1\% and 0.1\%).
We find that their performances are reduced by a significant amount, and are inferior to the model trained with pseudo-labels from AnyChange.
This confirms the potential of AnyChange as a change data engine for supervised object change detection. 
\vspace{-1mm}

\textbf{Unsupervised Change Detection Results.}
AnyChange's class-agnostic change masks are natural pseudo-labels for unsupervised change detection.
We also compare our AnyChange with unsupervised change detection methods.
In Table~\ref{tab:sota_second}, we find that AnyChange with ViT-B in a zero-shot setting improves over the previous state-of-the-art method, I3PE \citep{I3PE}.
To learn the biases on the SECOND dataset, we trained a ChangeStar (1$\times$256) model with pseudo-labels of AnyChange on the SECOND training set.
The setting follows I3PE \citep{I3PE}, thus we align their backbone and use ResNet50 for ChangeStar (1$\times$256).
The results show that two variants of unsupervised ChangeStar (1$\times$256) outperform I3PE. 
\begin{wraptable}{r}{0.5\textwidth}
\vspace{-4.2mm}
  \caption{\textbf{Unsupervised Change Detection}. Comparison with the state-of-the-art unsupervised change detectors on the \textbf{SECOND} \texttt{test}.
  ``*'' indicates this change detection model is trained with pseudo labels predicted by AnyChange.
    \label{tab:sota_second}}
  \centering
  \small
  \tablestyle{3pt}{1.2}
  \resizebox{\linewidth}{!}{
  \begin{tabular}{l|c|ccc}
    \toprule
     Method             & Backbone  & F$_1\uparrow$         & Prec. & Rec.  \\
     \hline
     ISFA  \citep{ISFA}  & -         & 32.9          & 29.8  & 36.8           \\
     DSFA \citep{DSFA}   & -         & 33.0          & 24.2  & 51.9           \\
     DCAE \citep{DCAE}   & -         & 33.4          & 35.7  & 31.4           \\
     OBCD \citep{OBCD}   & -         & 34.3          & 29.6  & 40.7           \\
     IRMAD \citep{IRMAD} & -         & 34.5          & 28.6  & 43.6           \\
     KPCA-MNet \citep{KPCAMNet} & -  &	36.7          & 29.5  & 48.5           \\
     DCVA \citep{DCVA}   & -         & 36.8          & 29.6  & 48.7           \\
     DINOv2+CVA (zero-shot, our impl.) & ViT-G/14  & 41.4          & 26.9  & \textbf{89.4}            \\
     I3PE \citep{I3PE}   & R-50      & 43.8          & \textbf{36.3}  & 55.3           \\
     \hline
     AnyChange (zero-shot, ours)      & ViT-H & 41.8       & 27.4  & 88.7    \\
     + ChangeStar (1$\times$256)* (ours) & R-50  & 45.0    & 30.2  & 88.2    \\
     AnyChange (zero-shot, ours)      & ViT-B & 44.6       & 30.5  & 83.2    \\
     + ChangeStar (1$\times$256)* (ours) & R-50  & \textbf{48.2}    & 33.5  & 86.4    \\
    \bottomrule
  \end{tabular}
  }
  \vspace{-3mm}
\end{wraptable}
Notably, based on pseudo-labels of AnyChange with ViT-B, our model set a new record of 48.2\% F$_1$ on the SECOND dataset for unsupervised change detection.
Besides, we obtain some useful insights for unsupervised change detection: (\textit{i}) deep features from VFMs significantly assist unsupervised change detection models since DINOv2+CVA beats all advanced competitors except I3PE. 
Before our strong baseline, DINOv2+CVA, CVA has been always regarded as a simple and ineffective baseline for unsupervised change detection.
(\textit{ii}) dataset biases are helpful for in-domain unsupervised change detection since our model trained with pseudo-labels achieves higher performance than these pseudo-labels.
This indicates the model learns some biases on the SECOND dataset, which may be change types and style.
\vspace{-4mm}

\section{Conclusion}
\label{conc}\vspace{-3mm}
We present the segment any change models (AnyChange), a new type of change detection model for zero-shot change detection, allowing fully automatic, semi-automatic with custom threshold, and interactive mode with simple clicks.
The foundation of all these capabilities is the intra-image and inter-image semantic similarities in SAM’s latent space we identified on multi-temporal remote sensing images.
Apart from zero-shot change detection, we also demonstrated the potential of AnyChange as the change data engine and demonstrated its superiority in unsupervised and supervised change detection.
AnyChange is an out-of-the-box zero-shot change detection model, and a step forward towards a ``foundation model'' for the Earth vision community.

\section*{Acknowledgements}
This work was supported in part by ARO (W911NF-21-1-0125), ONR (N00014-23-1-2159), the CZ Biohub, and the National Natural Science Foundation of China under Grant No. 42325105.

\medskip
{\small
\bibliographystyle{iclr2021_conference}
\bibliography{main}
}

\newpage
\appendix

\section{Appendix / supplemental material}
\section{Pseudocode of Bitemporal Latent Matching}
\label{supp:pseudocode}

\newcommand{\z}{\mathbf{z}}
\newcommand{\m}{\mathbf{m}}
\newcommand{\x}{\mathbf{x}}
\begin{algorithm}[ht]
  \caption{Bitemporal Latent Matching (top-k)\label{alg:blm}}
  \algrenewcommand\algorithmicindent{.5em}
  \newcommand{\HS}{\hspace{1em}}
  \begin{algorithmic}[1]
      \Require $\z_t, \z_{t+1}$: image embeddings; $\mathcal{M}_{t}, \mathcal{M}_{t+1}$: object proposals; $k$: number of change proposals
      \For{$i=1$ {\bf to} $N_t$}
        \State $\m_{t,i} \gets \mathcal{M}_{t}[i]$
        \State $\x_{t,i} \gets \z_t[\m_{t,i}]$
        \State $\hat{\x}_{t+1,i} \gets \z_{t+1}[\m_{t,i}]$
        \State $c_{t, i} \gets -d_m^{-1}\x_{t,i}\cdot\hat{\x}_{t+1,i}$ \HS $\triangleright$ compute change confidence score via embedding dissimilarity
      \EndFor
      \For{$j=1$ {\bf to} $N_{t+1}$}
        \State $\m_{t+1,j} \gets \mathcal{M}_{t+1}[j]$
        \State $\x_{t+1,j} \gets \z_{t+1}[\m_{t+1,j}]$
        \State $\hat{\x}_{t,j} \gets \z_{t}[\m_{t+1,j}]$
        \State $c_{t+1,j} \gets -d_m^{-1}\x_{t+1,j}\cdot\hat{\x}_{t,j}$ \HS $\triangleright$ compute change confidence score via embedding dissimilarity
      \EndFor
      \State $\{\m_i\}^k_{1} \gets$\texttt{sort}($\mathcal{M}_{t}\cup\mathcal{M}_{t+1}$, \texttt{key=lambda t,i:}$\{c_{t,i}\}\cup\{c_{t+1,j}\}[\texttt{t,i}]$)[:$k$] \HS $\triangleright$ Python-style top-k sorting
      \State {\bf return} $\{\m_i\}^k_{1}$
  \end{algorithmic}
\end{algorithm}
\vspace{-3mm}
\section{Implementation Details}

\subsection{Baselines for zero-shot change proposal and detection}
\label{supp:baseimpl}
\textbf{SAM forward:}
For object proposal generation, we adopt a point per side of 64, an NMS threshold of 0.7, a predicted IoU threshold of 0.5, and a stability score threshold is 0.8 for LEVIR-CD, S2Looking, SECOND, and 0.95 for xView2.
To obtain the image embedding with a constant $\ell_2$ norm ($\sqrt{d_m}$), we demodulate the output of the image encoder $f$ of SAM with the affine transformation ($\mathbf{w};\mathbf{b}$) of the last layer normalization, i.e., $\mathbf{z} = \mathbf{w}^{-1}(f(\mathbf{x}) - \mathbf{b})$, where $\mathbf{x}$ is the single input image.
The forward computation was conducted on 8 NVIDIA A4000 GPUs.

\textbf{DINOv2 + CVA:}
The implementation is straightforward where the image embedding is first extracted from DINOv2 and then upsampled into the image size with bilinear interpolation.
The change intensity map is computed by the $\ell_2$ norm of the difference between bitemporal image embeddings.
The optimal threshold is obtained by an adaptive threshold selection method, OTSU \citep{otsu1979threshold}.
We also conduct a linear search for its optimal threshold on a small validation set, and the searched results have comparable performance with OTSU.
Considering optimal peak performance and ease of use, we choose OTSU's threshold as the default for this baseline.

\textbf{SAM + Mask Match:}
The main idea of this baseline is to use the geometric difference to measure an object's change.
In general, if an object disappears at the next time image, the object mask generated by SAM is empty or has a different geometric shape, and vice versa.
Therefore, using their geometric shape difference to measure if the change occurred is reasonable.
To this end, we compute pairwise mask IoU between pre-event object masks and post-event object masks to match a potentially identical object at another time for each object.
We recognize the object region belongs to non-change when successfully matched (IoU$>$0.5), otherwise this region belongs to a change.

\textbf{SAM + CVA Match:}
This baseline follows the same idea of AnyChange, i.e., latent-based matching, but adopts the negative $\ell_2$ norm of feature difference as the similarity.
This method first computes pixel-level change map via SAM feature-based CVA, the procedure of which is similar to ``DINOv2 + CVA''.
The instance-level voting with a threshold of 0.5 is then adopted to obtain change proposals, which means that each region is considered as a change when more than half of the pixels are identified as changes.

\textbf{AnyChange:}
For a fair comparison and automatic benchmark evaluation, we use fully automatic mode for AnyChange in Table~\ref{tab:benchmark:zsocp}.
The change angle threshold (155$^{\circ}$) is obtained by OTSU and a linear search on the small validation set sampled from the SECOND training set.
This threshold is directly used for the other three datasets without any adjustment.

\textbf{AnyChange (Oracle):}
We only train a LoRA $(r=32, alpha=320, dropout=0.1)$ for the SAM model on each dataset due to unaffordable full-parameter training.
Besides, we attach a change confidence score network on the image encoder of SAM to predict an accurate semantic similarity instead of our negative cosine similarity.
This network architecture is composed of an MLP block (Linear-LayerNorm-GELU-Linear-LayerNorm), an upsampling block (ConvTranspose-LayerNorm-GELU-ConvTranspose-Linear-LayerNorm-Linear-LayerNorm) for 4$\times$ upsampling, and a linear layer for predicting change confidence score.
The loss function is a compound of binary cross-entropy loss and soft dice loss.
The inference pipeline exactly follows AnyChange, where it first generates object masks and computes mask embeddings, and the change confidence score is obtained from the trained score network.
The training iterations are 200 epochs with a batch size of 16, AdamW optimizer with a weight decay of 0.01.
The learning rate schedule is ``poly'' ($\gamma=0.9$) decay with an initial learning rate of 6e-5
The training data augmentation adopts random rotation, flip, scale jitter, and cropping.
The crop size is 512 for LEVIR-CD and S2Looking and 256 for xView2 and SECOND, respectively.

\subsection{Pseudo-label Training}
\label{supp:psetrain}
We have the training step of 20k, a batch size of 16, SGD with a momentum of 0.9, and a weight decay of 1e-3 for optimization on the SECOND training set with pseudo-labels of AnyChange in fully automatic mode.
For regularization, we adopt the label smoothing of 0.4 and strong data augmentations which include random crop to 256$\times$256, flip, rotation, scale jitter, and temporal-wise adjustment of brightness and contrast.

\section{Limitations}
\label{supp:lim}
Zero-shot change detection is an open problem in remote sensing and computer vision communities.
There is little reference to point out any potential effective roadmap before our work.
Our work is the first to define this problem suitably and provide a simple yet effective model and evaluation protocol.
The problem formulation and evaluation protocol themselves bring some potential limitations (i.e., scenario coverage, the robustness to objects of different geometries) since there is little mature infrastructure, e.g., a concept-complete class-agnostic change detection dataset.

\section{Broader Impacts}
\label{supp:broader}
The proposed method can detect class-agnostic changes in Earth's surface, however, its actual effect is impacted by SAM's latent space.
The model may produce some impossible changes due to SAM's biases.
These issues warrant further research and consideration when building upon this work for real-world applications.

\section{More visualization}

\begin{figure}[t]
\centering
\includegraphics[width=0.9\linewidth]{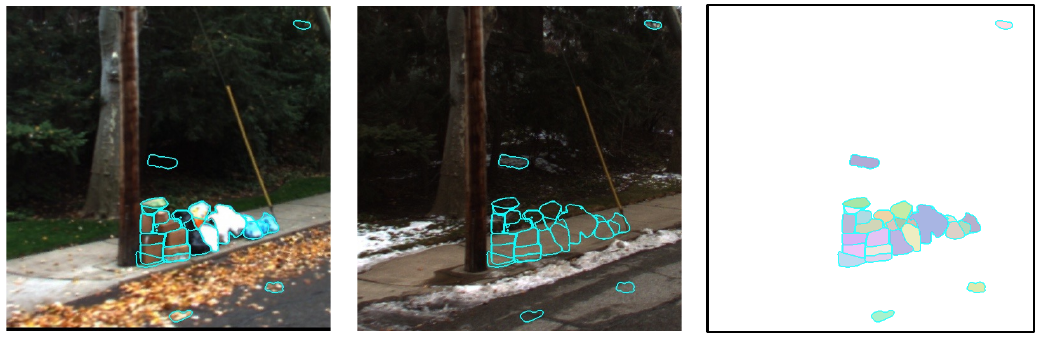}
\vspace{-3mm}
\caption{AnyChange on \textbf{Natural Image Domain}.
Best viewed digitally with zoom. 
}
\label{fig:natural_domain}
\end{figure}

\textbf{Nature Image Domain.}
Our method can be also used in the natural image domain to detect more general object changes, as shown in Fig.~\ref{fig:natural_domain}.

\textbf{More demonstration of point query mechanism.}
We provide an additional example to supplement Fig.~\ref{fig:pq}, which includes unchanged and changed buildings simultaneously. 
It is more clear to show that AnyChange can more accurately detect building changes with the help of the point query.

\begin{figure}[!t]
\centering
\includegraphics[width=0.9\linewidth]{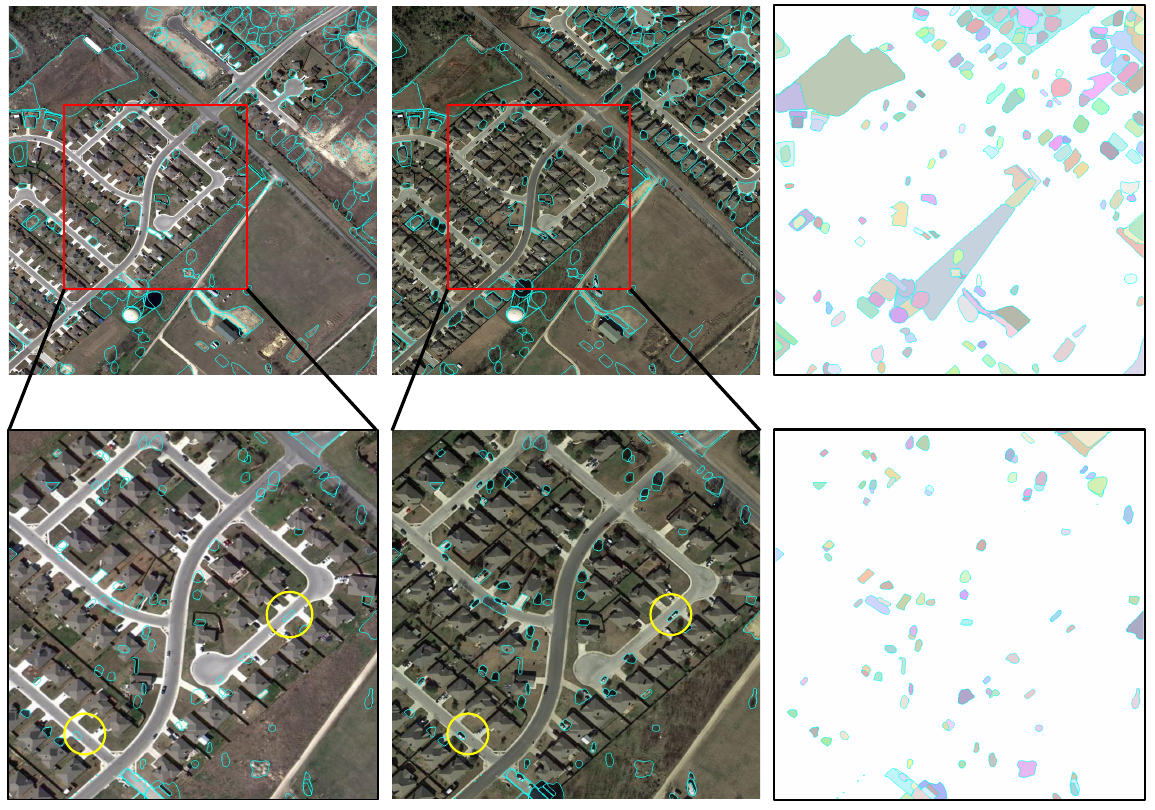}
\vspace{-3mm}
\caption{Illustration of the effectiveness of AnyChange in detecting tiny or minor object changes.
}
\label{fig:tinychange}
\end{figure}

\textbf{Effectiveness of AnyChange in detecting tiny object changes.}
Fig.~\ref{fig:tinychange} demonstrates a case of tiny/minor changes, such as small vehicle changes.
We observe that directly applying AnyChange to the original image overlooks these subtle changes (see the first row). 
After we bilinearly upsampled the red box region by 2$\times$ and then applied AnyChange to it, we find some tiny/minor changes could be detected (see the second row). 
This observation shows that our method has the ability on tiny object change detection, although it is not yet optimal. 
Future work could use our approach as a strong baseline to further improve the detection of subtle changes.

\begin{figure}[!t]
\centering
\includegraphics[width=0.9\linewidth]{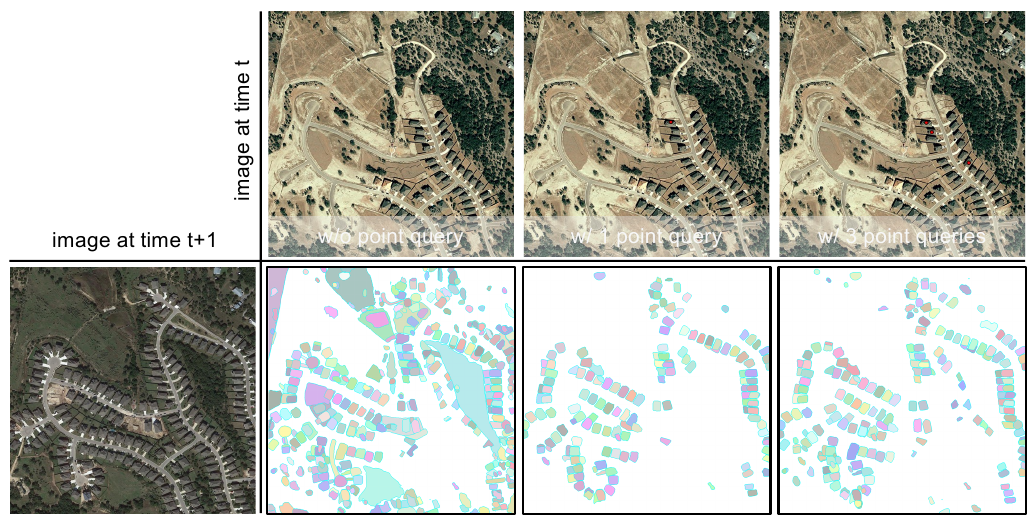}
\vspace{-3mm}
\caption{Examples of \textbf{Point Query Mechanism}.
The effects of w/o point query, one-point query, and three-point queries are shown in sequence from left to right.
(best viewed digitally with zoom, especially for the \textcolor{red}{red} points)
}
\label{fig:more_pq}
\end{figure}

\end{document}